# A Lightweight Approach for On-the-Fly Reflectance Estimation

Kihwan Kim[1] Jinwei Gu[1] Stephen Tyree[1] Pavlo Molchanov[1] Matthias Nießner[2] Jan Kautz[1]
[1]NVIDIA Research [2]Technical University of Munich
https://research.nvidia.com/publication/reflectance-estimation-fly

## Abstract

*Estimating surface reflectance (BRDF) is one key component for complete 3D scene capture, with wide applications in virtual reality, augmented reality, and human computer interaction. Prior work is either limited to controlled environments (e.g., gonioreflectometers, light stages, or multi-camera domes), or requires the joint optimization of shape, illumination, and reflectance, which is often computationally too expensive (e.g., hours of running time) for real-time applications. Moreover, most prior work requires HDR images as input which further complicates the capture process. In this paper, we propose a lightweight approach for surface reflectance estimation directly from 8-bit RGB images in real-time, which can be easily plugged into any 3D scanning-and-fusion system with a commodity RGBD sensor. Our method is learning-based, with an inference time of less than 90ms per scene and a model size of less than 340K bytes. We propose two novel network architectures, HemiCNN and Grouplet, to deal with the unstructured input data from multiple viewpoints under unknown illumination. We further design a loss function to resolve the color-constancy and scale ambiguity. In addition, we have created a large synthetic dataset, SynBRDF, which comprises a total of 500K RGBD images rendered with a physically-based ray tracer under a variety of natural illumination, covering 5000 materials and 5000 shapes. SynBRDF is the first large-scale benchmark dataset for reflectance estimation. Experiments on both synthetic data and real data show that the proposed method effectively recovers surface reflectance, and outperforms prior work for reflectance estimation in uncontrolled environments.*

## 1. Introduction

Capturing scene properties in the wild, including its 3D geometry and surface reflectance, is one of the ultimate goals of computer vision, with wide applications in virtual reality, augmented reality, and human computer interaction. While 3D geometry recovery has achieved high accuracy — especially given recent RGBD-based scanning-and-fusion approaches [7, 22, 33], surface reflectance estimation, however, still remains challenging. At one extreme, most of the fusion methods assume Lambertian reflectance and recover surface texture only. At the other extreme, most of the prior work on surface BRDF (bidirectional reflectance distribution function) estimation [14, 20, 19] aims to recover the full 4D BRDF function, but are often limited to controlled, studio-like environments (e.g., gonioreflectometers, light stages, multi-camera domes, planar samples).

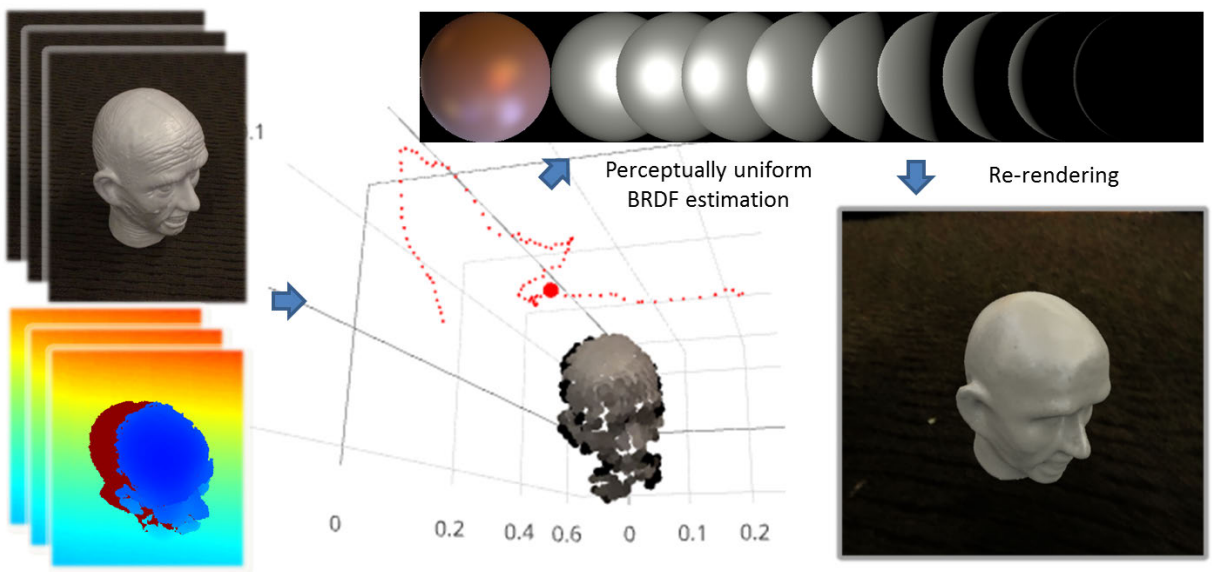

(a) (b) (c)

Figure 1. Overview of our method: we take RGBD image sequences as inputs (a). During a reconstruction process, each view contributes voxels as an observation. In (b), colors of visible voxel samples covered by a specific view (red circle) are visualized (red dots are the locations of other views). These samples from all views are evaluated through either HemiCNN (Sec. 3.3.1) or Grouplet networks (Sec. 3.3.2) to estimate the BRDF in real time. In (c), we show a rendered image under a novel viewing and lighting condition. More examples are shown in Sec. 4.

Recently, a few methods [35, 18, 27, 26, 30, 17, 1] were proposed to recover surface reflectance in uncontrolled environments (e.g., unknown illumination or shape) by utilizing statistical priors on natural illumination and/or BRDF. These methods formulate the inverse rendering problem as a joint, alternative optimization among shape, reflectance, and/or lighting. Despite their accuracy, these methods are computationally quite expensive (e.g., hours or days of running time and tens of gigabytes of memory consumption) — they are often run in a post-process rather than a real-time setting. Moreover, these methods often require high-resolution, HDR images as input, which further complicates the capturing process for real-time or interactive applications on consumer-grade mobile devices.

In this paper, we propose a lightweight and practical approach for surface reflectance estimation directly from 8-bit RGB images in real-time, which can be easily plugged into any 3D scanning-and-fusion system with a commodity RGBD sensor, and enables physically-plausible renderings at novel lighting/viewing conditions, as shown in Fig. 1. Similar to prior work [35, 18], we use a simplified BRDF representation and focus on estimating surface albedo and gloss rather than full 4D BRDF function. Our method is learning-based, with an inference time of less than 90ms per scene and a model size of less than 340K bytes.

In order to deal with unstructured input data (e.g., each surface point can have a different number of observations due to occlusions) in the context of neural networks, we propose two novel network architectures – HemiCNN and Grouplet. HemiCNN projects and interpolates the sparse observations onto a 2D image, which enables the use of standard convolutional neural networks. Grouplet learns directly from random samples of observations and uses multi-layer perception networks. Both networks are also designed to be lightweight in both inference time and model size for real-time applications on consumer-grade mobile devices. In addition, since the illumination is unknown, we have designed a novel loss function to resolve the color-constancy and scale ambiguity (i.e., only given input images, we do not know whether surfaces are reddish or the lighting is reddish, or whether surfaces are dark or the lighting is dim). *To the best of our knowledge, this is the first lightweight, real-time approach for surface reflectance estimation in the wild.*

We also created, a large-scale synthetic dataset – SynBRDF – for reflectance estimation. SynBRDF covers 5000 materials randomly sampled from OpenSurfaces [3], 5000 shapes, and in total 500K RGBD images (both HDR and LDR) rendered from multiple viewpoints with a physically-based ray tracer under 20 natural environmental illumination conditions, which makes it an ideal benchmark dataset for complete image-based 3D scene reconstruction.

Finally, we have incorporated the proposed method in the RGBD scanning-and-fusion method for complete 3D scene capture (see Sec. 4 and Fig. 8). We trained our networks with SynBRDF, and directly applied the trained models on real data captured with a commodity RGBD sensor. Experiments on both synthetic data and real data show that the proposed method effectively recover surface reflectance, and outperforms prior work for surface reflectance estimation in uncontrolled environments.

## 2. Related Work

**Surface Reflectance Estimation in Uncontrolled Environments** Most of prior work in this direction formulate the inverse rendering problem as a joint optimization among the three radiometric ingredients — lighting, geometry, and reflectance — from observed images. Barron et al. [1] assumes Lambertian surfaces and optimizes all the three components. Others [30, 8, 17, 9] optimize reflectance and illumination with known 3D geometry, either from motion or based on statistical priors on natural illumination and materials. Recently, Wu et al. [35] and Lombardi et al. [18] proposed to jointly estimate lighting, reflectance and 3D shape from a RGB-D sensor, even in the presence of interreflection. Chandraker et al. [5] investigated the theoretical limits of material estimation from a single image. Despite the effectiveness, these methods solve complicated optimization problems iteratively, which are computationally quite expensive for real-time applications (e.g., hours of running time). As the optimization relies heavily on the parametric forms of statistical priors, these methods generally require good initialization and HDR images as input. In contrast, our proposed method is a lightweight and practical approach that can estimate surface reflectance directly from 8-bit RGB images on-the-fly, which is suitable for real-time applications.

**Material Perception and Recognition** Our work is also inspired from prior work on material perception and recognition from images. Pellacini et al. [28] designed a perceptually-uniform representation of the isotropic Ward BRDF model [32], and Wills et al. [34] extends to data-driven models with measured reflectance. Fores et al. [11] studies the metrics used for BRDF fitting [23]. Fleming et al. [10] found that natural illumination is key for the perception of surface reflectance. Bell et al. [3] released a large dataset OpenSurfaces, with annotated surface appearance from real-world materials. These prior works inspires us for designing the regression loss and creating the synthetic dataset for training. For learning-based material recognition, Liu et al. [16] proposed a Bayesian approach based on a bag of visual features. Bell et al. [4] used CNNs (convolutional neural networks) for material recognition from material context input. Recently, Wang et al. [31] proposed a CNN-based method for material recognition from light field images. These prior work shows neural network is capable of learning discriminative features for material perception from images.

**Reflectance Maps Estimation and Intrinsic Image Decomposition** Intrinsic image decomposition aims to factor an input image into a shading-only image and a reflectance-only image. Recently, CNNs has been successfully employed for intrinsic image decomposition [15, 21] from a single image. Bell et al. [2] proposed a dense CRF-based method and released a large intrinsic image dataset by crowdsourcing. Zhou et al. [36] used deep learning to infer data-driven priors from sparse human annotations. Rematas et al. [29] used CNNs to estimate reflectance maps (i.e., defined as 2D reflectance under the fixed, unknown illumination) from a single image. These methods recover only the illumination-dependent reflectance map, while our method estimates the full BRDF that enables rendering under novel illumination and viewing conditions.

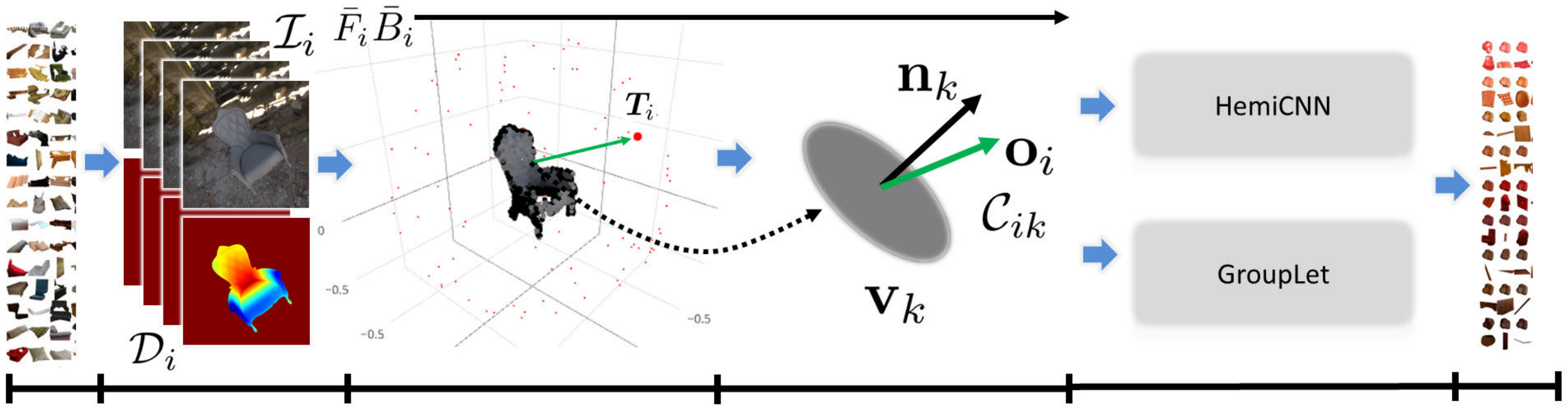

Figure 2. **Overview of our framework: (a)** BRDF examples from OpenSurface [3], **(b)** Input image $\mathcal{I}_i$ and depth $\mathcal{D}_i$ streams. Integrated volume is shown in **(c)**, where the colors shown from $i$th view $\mathcal{I}_i$ (the red circle with pose $\boldsymbol{T}_i$) are visualized. Small dots refer to the locations of other views. **(d)** shows the data that we extract from each voxel $\mathbf{v}_k$ for training; normal $\mathbf{n}_k$, observation vector $\mathbf{o}_i$ (the green arrow), color values $\mathcal{C}_{ik}$ from the observation $\mathcal{I}_i$ at the voxel $\mathbf{v}_k$. In **(e)**, these measurements together with color statistics ($\bar{F}_i$ and $\bar{B}_i$) are fed into one of the two networks, HemiCNN (Sec. 3.3.1) and Grouplet (Sec. 3.3.2) for BRDF estimation.

## 3. Method Overview

Our goal is to develop a module for real-time surface reflectance estimation that can be plugged into any 3D scanning-and-fusion methods for 3D scene capture, with potential applications in VR/AR. In this paper, we make a significant step towards this goal, and propose two novel networks for *homogeneous* surface reflectance estimation from RGBD image input.

### 3.1. Framework and Reflectance Model

Our framework takes as input RGBD image/depth sequences from a commodity depth sensor (Fig. 1). We denote RGB color observation as $\mathcal{C} : \Omega_{\mathcal{C}} \to \mathbb{R}^3$, images with $\mathcal{I} : \Omega_{\mathcal{I}} \to \mathbb{R}^3$ and depth maps with $\mathcal{D} : \Omega_{\mathcal{D}} \to \mathbb{R}$. The $N$ acquired RGBD frames consist of RGB color images $\mathcal{I}_i$, and depth maps $\mathcal{D}_i$ (with frame ID, $i \in 1 \dots N$). We also denote the absolute camera poses $\boldsymbol{T}_i = (\boldsymbol{R}, \boldsymbol{t}) \in \mathrm{SE}(3)$, $\boldsymbol{t} \in \mathbb{R}^3$ and $\boldsymbol{R} \in \mathrm{SO}(3)$ of the respective frames, which is computed from standard volume-based pose-estimation algorithm [7].[1] As shown in Fig. 2, the input $\mathcal{I}_i$, and $\mathcal{D}_i$ are aligned and integrated into a 3D volume $V$ with signed-distance fusion [24], from which, we extract voxels $\mathbf{v}_k \in V$ ($k \in 1 \dots M$) that contain observed color $\mathcal{C}_{ik}$ from the corresponding view $\mathcal{I}_i$, its surface normal $\mathbf{n}_k$, and camera orientation $\mathbf{o}_i$, see Fig. 2(d). Additionally, we compute the color statistics for each view by simply taking the average of foreground and background pixels in $\mathcal{I}_i$. We denote these as $\bar{F}_i$ and $\bar{B}_i$, which we will discuss further in Sec. 3.2.

For the representation of surface reflectance, similar to prior work [35, 18], we chose a parametric BRDF model – the isotropic Ward BRDF model [32] – due to two reasons: (1) the Ward BRDF model has a simple form but is representative for a wide variety of real-world materials [23], and (2) prior studies [28, 34] on BRDF perception are based on the Ward BRDF model. Specifically, the isotropic Ward BRDF model is given by:

$$f(\omega_i, \omega_o; \Theta) = \frac{\rho_d}{\pi} + \rho_s \cdot \frac{\exp\left(-\tan^2\theta_h/\alpha^2\right)}{4\pi\alpha^2\sqrt{\cos\theta_i\cos\theta_o}}, \tag{1}$$

where $\omega_i = (\theta_i, \phi_i)$ and $\omega_o = (\theta_o, \phi_o)$ are the incident and viewing directions, $\theta_h$ is the half angle, and $\Theta = (\rho_d, \rho_s, \alpha)$ is the parameters to be estimated.

An equivalent, but perceptually-uniform representation of the Ward BRDF model was proposed in [28], where the diffuse albedo $\rho_d$ is converted from RGB to CIE Lab colorspace, and the gloss is described by two variables — $c$, the contrast of gloss, and $d$, the distinctness of gloss. These two variables $c$ and $d$ are related to the BRDF parameters as follows [28]:

$$c = \sqrt[3]{\rho_s + \rho_d/2} - \sqrt[3]{\rho_d/2}, \quad d = 1 - \alpha. \tag{2}$$

Thus, an alternative representation for the BRDF parameters is $\Theta = (L, a, b, c, d)$.

Our problem is thus formulated as follows. *Given a set of voxels from any 3D scanning-and-fusion pipeline,* $\{\mathbf{v}_k\} = \{\{\mathcal{C}_{ik}, \mathbf{o}_i\}, \mathbf{n}_k\}$, *we estimate the optimal BRDF parameters* $\Theta$ *with neural networks.* Two problems need to be solved for learning. First, what is a good loss function that can resolve the color constancy and scale ambiguities due to unknown illumination? For example, just from input images, we cannot tell whether the material is reddish or the illumination is reddish, or whether the material is dark or the illumination is dim. Second, the input data is unstructured — different voxels have different numbers of observations due to occlusion. What is a good network architecture for such unstructured input data? We address these two problems in the following sections.

[1] During training, we randomly generated poses for rendering scenes.

| Name | $\mathbf{E_d}(\mathbf{\Theta}, \hat{\mathbf{\Theta}})$ |
|---|---|
| $\text{RMSE}_1$ | $\|\|\rho_d - \hat{\rho_d}\|\|^2 + \|\|\rho_s - \hat{\rho_s}\|\|^2 + \|\|\alpha - \hat{\alpha}\|\|^2$ |
| $\text{RMSE}_2$ | $\|\|\text{Lab} - \hat{\text{Lab}}\|\|^2 + \lambda_g \|\|\text{cd} - \hat{\text{cd}}\|\|^2$ |
| Cubic Root | $\sqrt[3]{\int_{\omega_i,\omega_o} \|\|f(\omega_i,\omega_o;\Theta) - f(\omega_i,\omega_o;\hat{\Theta})\|\| \cos\theta_i d\omega_o d\omega_o}$ |

Table 1. Three options for the distance function $E_d(\Theta, \hat{\Theta})$ for BRDF estimation. $\text{RMSE}_1$ is a Root Mean Squared Error of $\Theta = (\rho_d, \rho_s, \alpha)$, $\text{RMSE}_2$ is for $\Theta = (L, a, b, c, d)$, which is the sum of the perceptual color difference and the perceptual gloss difference ($\lambda_g = 1$) [28], and Cubic Root is the cosine-weighted $\ell_2$-norm of the difference of two 4D BRDF functions, inspired by BRDF fitting [23, 11].

### 3.2. Design of the Loss Function

A key part for network training is an appropriate loss function. Prior work [23, 11] showed that the commonly-used $\ell_2$ norm (i.e., MSE) is not optimal for BRDF fitting. We thus design the loss as

$$J = E_d(\Theta, \hat{\Theta}) + \lambda E_c(\hat{\Theta}, \{\mathcal{C}_{ik}\}) \tag{3}$$

where $E_d(\cdot,\cdot)$ is the data loss that measures the discrepancy between the estimate BRDF parameters and the ground truth, and $E_c(\cdot,\cdot)$ is the regularization term which relates the estimated reflectance $\hat{\Theta}$ with the observed image intensities $\{\mathcal{C}_{ik}\}$. $E_c$ is added to resolve the aforementioned scale ambiguity and color constance ambiguity. $\lambda$ is a weight coefficient, and $\lambda = 0.01$ in all our experiments.

Table (1) lists three options of $E_d(\Theta, \hat{\Theta})$ implemented in this paper. $\text{RMSE}_1$ is a Root Mean Squared Error of $\Theta = (\rho_d, \rho_s, \alpha)$, $\text{RMSE}_2$ is for $\Theta = (L, a, b, c, d)$, which is the sum of the perceptual color difference and the perceptual gloss difference ($\lambda_g = 1$) [28]. Cubic Root is inspired from BRDF fitting [23, 11] which is a cosine weighted $\ell_2$-norm of the difference between two 4D BRDF functions.

For $E_c$, we use the color statistics computed for each view, $\bar{F}_i$ and $\bar{B}_i$, to approximately constrain the estimation of $\rho_d$ and $\rho_s$. Specifically, $E_c$ is derived based on the rendering equation [13] as

$$E_c = \sum_i ||(\rho_d + \rho_s) \cdot \bar{B}_i^{\gamma} - \bar{F}_i^{\gamma}||^2, \tag{4}$$

where $\bar{F}_i$ and $\bar{B}_i$ are the average image intensity of the foreground and background regions of the $i$-th input image $I_i$, $\gamma = 2.4$ is used to convert the input 8-bit RGB images to linear images; see the appendix for a detailed derivation. Even though Eq. (4) is only an approximation of the rendering equation, it imposes a soft constraint on the scale and color cues for the BRDF estimation. We found it quite effective for working with real data (see Fig. 8).

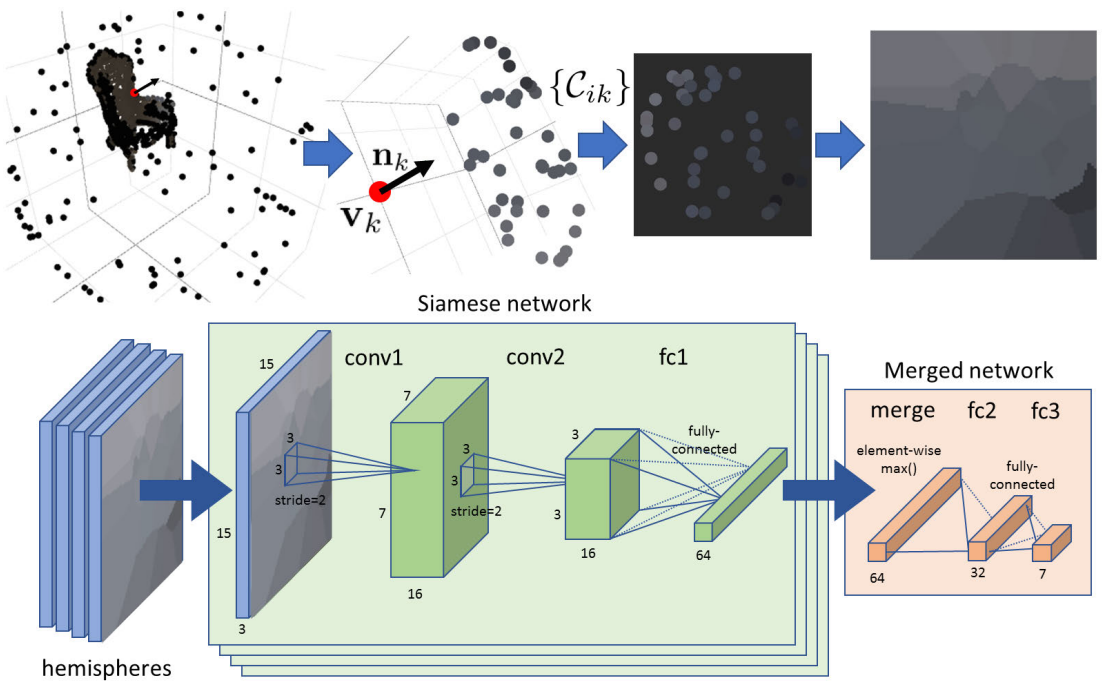

Figure 3. Details of HemiCNN. *Top row:* generating a voxel hemisphere image, from the sparse 3-D set of the observations $\{\mathcal{C}_{ik}\}$ of voxel $v_k$ to a dense 2-D hemisphere representation. *Bottom row:* the HemiCNN siamese convolutional neural network architecture.

### 3.3. Network Architectures

As shown in Fig. 2, our input data is unstructured, because the observations $\mathcal{C}_{ik}$ for each voxel $\mathbf{v}_k$ are irregular, sparse samples on the 2D slice of the 4D BRDF. Different voxels may have different numbers of observations due to occlusion. In order to feed the unstructured input data into networks for learning, we proposed two new neural network architectures. One is called Hemisphere-based CNN (HemiCNN) which projects and interpolates the sparse observations onto a 2D image that enables the use of standard convolutional neural networks. The other one is called Grouplet, which learns directly from randomly sampled observations and uses multilayer perceptron networks. Both networks are also designed to be lightweight in both inference time ($\leq$ 90ms) per scene and model size ($\leq$ 340K bytes) for real-time applications.

#### 3.3.1 Hemisphere-based CNN (HemiCNN)

Specifically, for HemiCNN, as shown in the top row of Fig. 3, the RGB observations $\{\mathcal{C}_{ik}\}$ at voxel $\mathbf{v}_k$ are projected onto a unit sphere centered at the sample voxel $\mathbf{v}_k$. The unit sphere is rotated so the positive $z$-axis is aligned with the voxel's surface normal $\mathbf{n}_k$. Observations $\{\mathcal{C}_{ik}\}$ on the positive hemisphere (i.e., $z > 0$) are projected onto the 2D $x$-$y$ plane. Finally, a dense 2D image, denoted as a sample hemisphere image, is generated using nearest-neighbor interpolation among the projected observations.

A siamese convolutional neural network is used to predict BRDF parameters from a collection of $N$ sample hemisphere images, one for each of a representative set of voxels; e.g., chosen by clustering on voxel positions or surface normals. As shown in Fig. 3, in the first of two stages, the siamese convolutional network operates on each sample hemisphere image individually to produce a vector representation, after which the representations are merged across the $N$ samples (i.e., voxels) by computing an element-wise maximum. The network includes two **conv** layers, each with 16 sets of $3 \times 3$ filters and ReLU activations, a single $2 \times 2$ max-pooling layer, and a single **fc** layer with 64 neurons. After aggregating the $N$ feature vectors with an elementwise-maximum, we use an **fc** layer with 32 neurons, followed by **tanh** activation, and then a final **fc** layer to produce the BRDF prediction. In most experiments with HemiCNN, we set $N = 25$. The model size is 56K bytes and the average inference time is 16ms per scene.

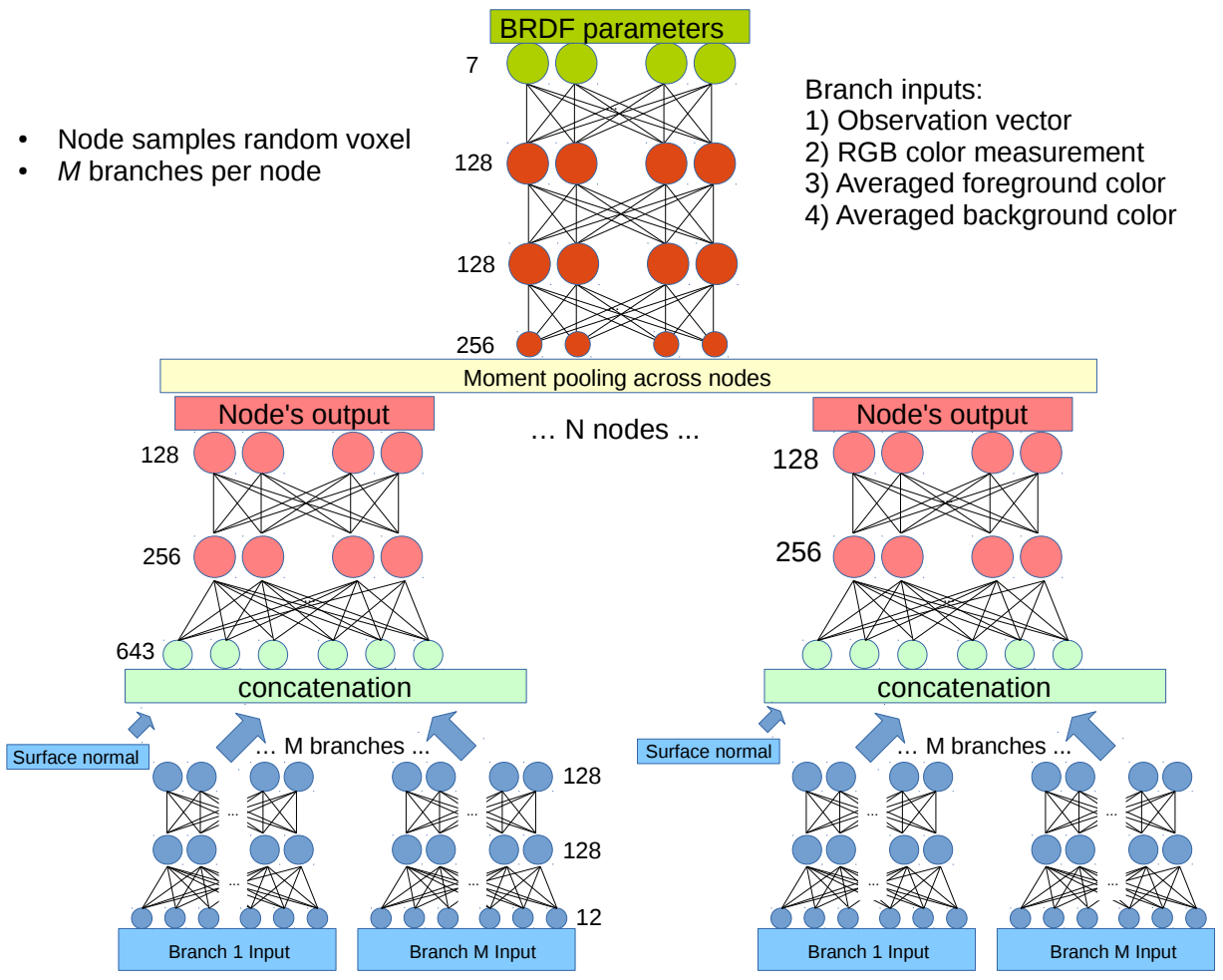

Figure 4. Grouplet for BRDF estimation. It relies on aggregating results from a set of weak regressors (nodes). Each node operates on a randomly sampled voxel from the object. M branches form input to the node; they are sampled randomly from a set of observations per voxel. Intermediate representation from multiple nodes is combined by moment pooling layer. Finally, BRDF parameters are regressed from the output of the moment pooling layer.

#### 3.3.2 Sampling-based Network (Grouplet)

The second proposed network architecture is called Grouplet (Fig. 4). Unlike HemiCNN, where we transform sparse observations to 2D images to use standard convolution layers, Grouplet directly operates on each observation $\mathcal{C}_{ik}$ for each voxel $\mathbf{v}_k$. Grouplet relies on aggregating results from a set of weak regressors called *nodes*. Each node estimates an intermediate representation of the BRDF parameters of a single voxel ($\mathbf{v}_k$) and randomly sampled $M$ observations $\Gamma = \{\mathcal{C}_{1k}, \cdots, \mathcal{C}_{Mk}\}$, as shown in Fig. 4. Subsets of all the observations $\Gamma$ are selected randomly per each voxel. Each observation from $\Gamma$ is processed by a two-layer multilayer perceptron (MLP) (with 128 neurons for each layer), called *branch*, that takes four inputs: observed color ($\mathcal{C}_{ik}$), viewing direction ($\mathbf{o}_i$), averaged foreground color ($\bar{F}_i$) and averaged background color ($\bar{B}_i$).

Next, we combine information from the $M$ branches by concatenating their outputs together with the voxel's surface normal ($\mathbf{n}_k$). We thus obtain a 643-dimensional representation. This vector is processed by another two-layer MLP with 256 and 128 neurons in the layers, the output of which is the intermediate representation of the BRDF parameters. During BRDF estimation, we operate on $N$ voxels, each of which is processed by different nodes with shared weights. To combine the intermediate representations computed from several voxels, we use a moment pooling operator that is invariant to the number of nodes. We pool with the first and second central moments which represent expected value and variance of the intermediate representation across nodes. The output of the pooling operator is a 256-dimensional pooled representation.

The final part of the network estimates the BRDF parameters from the pooled representation by another MLP with two hidden layers of 128 neurons each, and one final output layer. All layers in every branch, node and the final regressor have hyperbolic tangent activation functions, except for the last output layer.

Grouplet is flexible because of its capability to work with any number of nodes due to the usage of pooling operators. It also does not require that the number of nodes will be the same during testing and training time. The order of the $M$ observations is important for network to operate correctly. We found that arranging observations by the cosine distance between the observation vector ($\mathbf{o}_i$) and the voxel's surface normal ($\mathbf{n}_k$) leads to the best result. For BRDF estimation, Grouplet is applied in two forms: *Grouplet-fast* and *Grouplet-slow* that take $N = 20$ and $N = 354$ voxels respectively. Each of them construct $M = 10$ nodes per voxel. The average inference time for Grouplet-fast is 5 msec, and for Grouplet-slow is 90 msec. The model size for both Grouplet-fast and Grouplet-slow is 339K bytes. In this paper, we refer to Grouplet as the Grouplet-slow unless otherwise stated.

For both HemiCNN and GroupLet, we set $\lambda_g = 1$ and explore a range of $\lambda$, finding $0.1 \leq \lambda \leq 1$ to be a reasonable range. We train HemiCNN using RMSProp with learning rate 0.0001 and 100K minibatches. For Grouplet training, we use stochastic gradient descent with fixed learning rate 0.01 and momentum 0.9 for 13K minibatches.

### 3.4. SynBRDF: A Large Benchmark Dataset

Deep learning requires large amount of data. Yet, for BRDF estimation, it is extremely challenging to obtain a large dataset with measured BRDF data due to the complex settings for BRDF acquisition [19, 17]. Moreover, while there are quite a few recent work for 3D shape recovery and reflectance estimation in the wild [17, 27, 35], one can hardly find any large-scale, benchmark datasets with ground-truth shape, reflectance and illumination.

With these motivations, in this paper, we created, to our knowledge, the first large-scale, synthetic dataset, called SynBRDF. SynBRDF covers 5000 materials randomly sampled from OpenSurfaces [3], 5000 shapes randomly sam-pled from ScanNet [6], and in total 500K RGB and depth images (both HDR and LDR) rendered with a physically-based raytracer [12], under variants of 20 natural environmental illumination from multiple viewpoints. SynBRDF thus has groundtruth for 3D shape, BRDF, illumination, and camera poses, which makes itself ideal as a large benchmark dataset for evaluating image-based 3D scene reconstruction and BRDF estimation algorithms. As shown in Fig. 5, each scene is labeled with a ground truth Ward BRDF parameters (Eq. 1 and 2). For more flexible evaluation that allows other types of rendering (e.g., global illumination) for the same scene, we will also provide the 90K xml scripts that fully covers original OpenSurface materials, and contains the variations of environmental and object models settings. We believe this dataset will be valuable for further study in this direction.

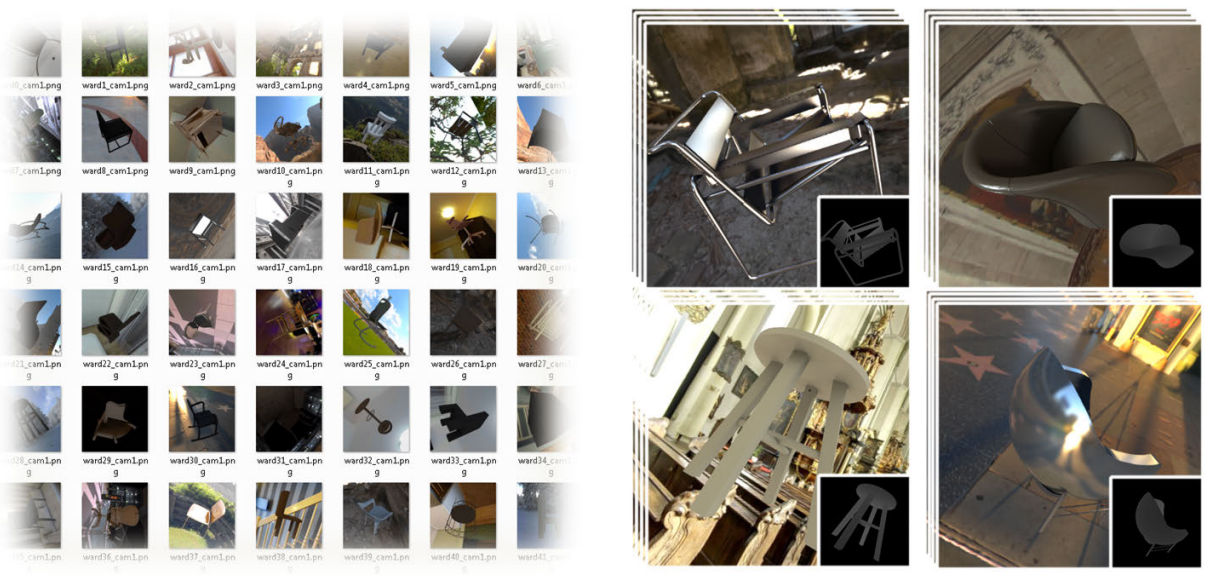

Figure 5. **SynBRDF**: (Left) Thumbnails of the first frame of each example (each contains 100 different observations), (Right) Some examples with depth map (insets)

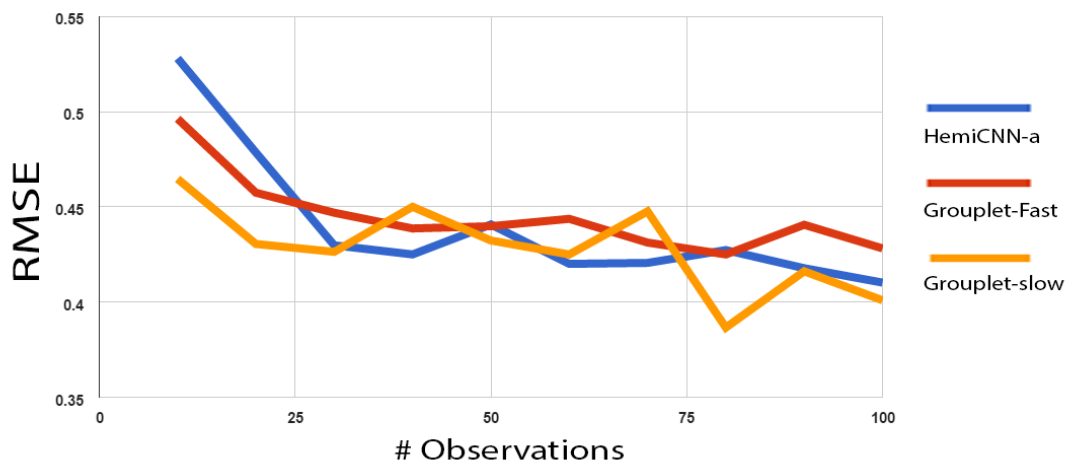

Figure 6. Coverage of observations and voxel samples: The blue line shows the RMSE from one of the HemiCNN settings, the red line denotes Grouplet-fast (20 voxels), and the orange line denotes the Grouplet-slow (354 voxels). This graph shows how much denser information needed for reasonable prediction. In fact, the common scenario of *scan-and-fusion* does not always guarantee the rich coverage of observations. We see the effective amount of coverage of views (observations) around 30. Similar tendency also happens for larger number of voxels.

In our experiments, we used SynBRDF for training and validation. We randomly chose 400 from the total 5000 scenes for validation, and the remaining 4600 scenes for training. For real data experiments, we directly applied the trained models without any domain adaptation.

| Network | Loss | RMSE | User Rank |
|---|---|---|---|
| Grouplet | $\text{RMSE}_1 + E_c$ | 0.455 | 1 |
| Grouplet | $\text{RMSE}_1$ | 0.432 | 2 |
| HemiCNN | $\text{RMSE}_2$ | 0.564 | 3 |
| HemiCNN | CubeRoot | 0.439 | 4 |
| HemiCNN | $\text{RMSE}_1$ | 0.419 | 5 |
| HemiCNN | $\text{CubeRoot} + E_c$ | 0.583 | 6 |
| Grouplet | $\text{RMSE}_2$ | 0.457 | 7 |

Table 2. Average RMSE (w.r.t ground truth BRDF parameters) on the validation set of SynBRDF and the rank of user preferences from a perceptual study of rendered materials. Among the variants of our proposed method, we list the top seven methods based on the results of the user study, and in general these provide most plausible results among all testing data (see results in Fig. 7 and Fig. 13). Note that RMSE ranking is not always consistent with the ranking of user study. As shown in Fig. 8, CubeRoot+$E_c$ provides most plausible results for the real data. Additional evaluations are found in the supplementary material

# 4. Experimental Results

We evaluated multiple variants of the proposed networks, changing the loss function $E_d$ and BRDF representations as described in Sec 3.1. In Sec. 4.1, we evaluate these settings on SynBRDF, showing quantitatively and qualitatively that several combinations give accurate predictions on our synthetic dataset. In Sec. 4.2, we compare with prior work [17]. Finally in Sec. 4.3, we demonstrate the proposed methods within the KinectFusion pipeline for complete 3D scene capture with real data.

## 4.1. Results on Synthetic Data

We evaluated all variants of the proposed methods on the validation set of SynBRDF. Evaluating the quality of BRDF estimation is challenging [11]—perceived quality often varies with the illumination, 3D shape, and even display settings. Estimates with the lowest RMSE error on the BRDF parameters are not necessarily the best for visual perception. Thus, in addition to computing the RMSE with respect to the ground truth BRDF parameters, we also conducted a user study. We randomly chose 10 materials from the validation set, and rendered the BRDF predictions for each material under (a) natural illumination and (b) moving point light sources. The rendered images are similar to Fig. 7. We then asked 10 users to rank the methods on each material based on the perceptual similarity between the ground truth and the images rendered from each method.

Table 2 lists the top seven methods based on average user score, together with the RMSE w.r.t ground truth BRDF parameters.[2] The max user score is 100, and the higher the

[2]RMSE is computed after normalization of the BRDF parameters to zero mean and unit standard deviation, based on the mean and standard

Material 947: **top**: GT, **middle: ours (RMSE$_1+E_c$)**, **bottom**: Lombardi et al.

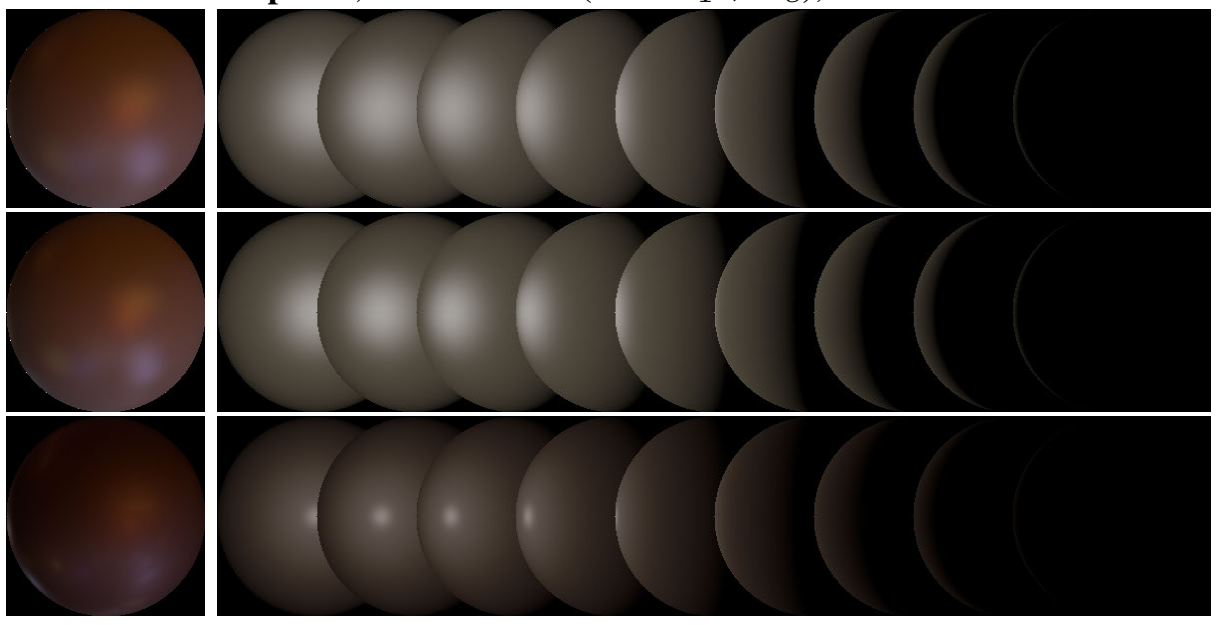

Material 3331: **top**: GT, **middle: ours (RMSE$_1$)**, **bottom**: Lombardi et al.

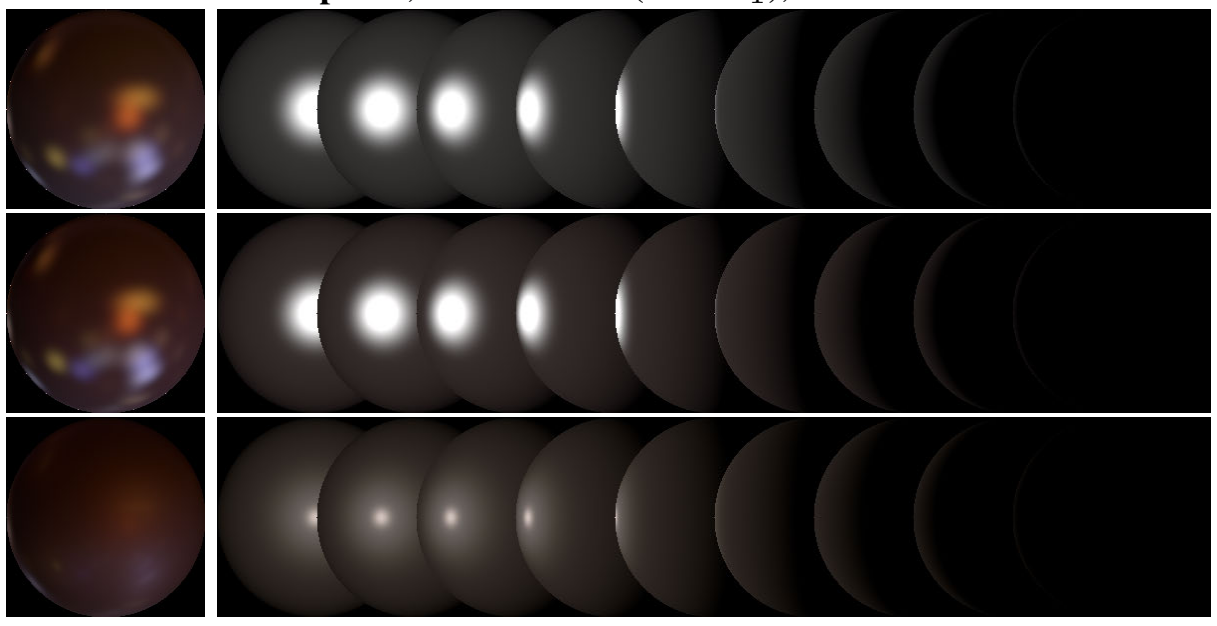

Figure 7. Comparison with Lombardi et al [17]: ours (middle row for each example) shows more closer to the ground truth surface examples even under various lighting.

better. As shown in Table 2, first, we found the RMSE ranking is not always consistent with the ranking of user study. In general, we found adding the regularization term $E_c$ improves the estimation (e.g., Grouplet-RMSE$_1$-$E_c$), while the choice of $E_d$ does not affect the performance much. Moreover, we found that the $\Theta = (L, a, b, c, d)$ BRDF representation provides more accurate estimation of gloss (e.g., HemiCNN-RMSE$_2$). Also, HemiCNN seems able to obtain better estimate for the gloss, while Grouplet estimates the diffuse albedo better. More complete evaluation results are provided in the supplementary material.

Fig. 13 shows three random examples of BRDF estimation results, where we show the rendered images under natural illumination with the ground truth BRDFs, as well as the estimated BRDFs from two variants of our proposed methods that have lowest RMSE. Qualitatively, the BRDF estimations accurately reproduce the color and gloss of the surface materials.

## 4.2. Comparison with [17]

As mentioned previously, it is difficult to compare with prior work on BRDF estimation in the wild [27, 35], given the lack of code and common datasets for comparison. Lombardi et al. [17] is the only method with released codes. Strictly speaking, it is not a direct apple-to-apple comparison, because Lombardi et al. [17] requires a single image and a precise surface normal map as input and estimates both DSBRDF and lighting, while our methods take multiple RGB-D images as input and estimate the Ward BRDF. Moreover, Lombardi et al. [17] takes about 3 minutes to run, while our methods are real-time ($\leq$ 90ms). Nevertheless, Lombardi et al. [17] is the only available option for comparison, and both its input requirements and running time are similar to ours. For comparison, specifically, we randomly choose materials from SynBRDF, rendered a sphere image under natural illumination, and used it (together with the sphere normal map) as the input for [17]. Fig. 7 shows the comparison results for two materials. It shows our proposed method outperforms [17].

deviation of the training set. Thus a random prediction (with the same statistics) will have RMSE $\approx$ 1.0.

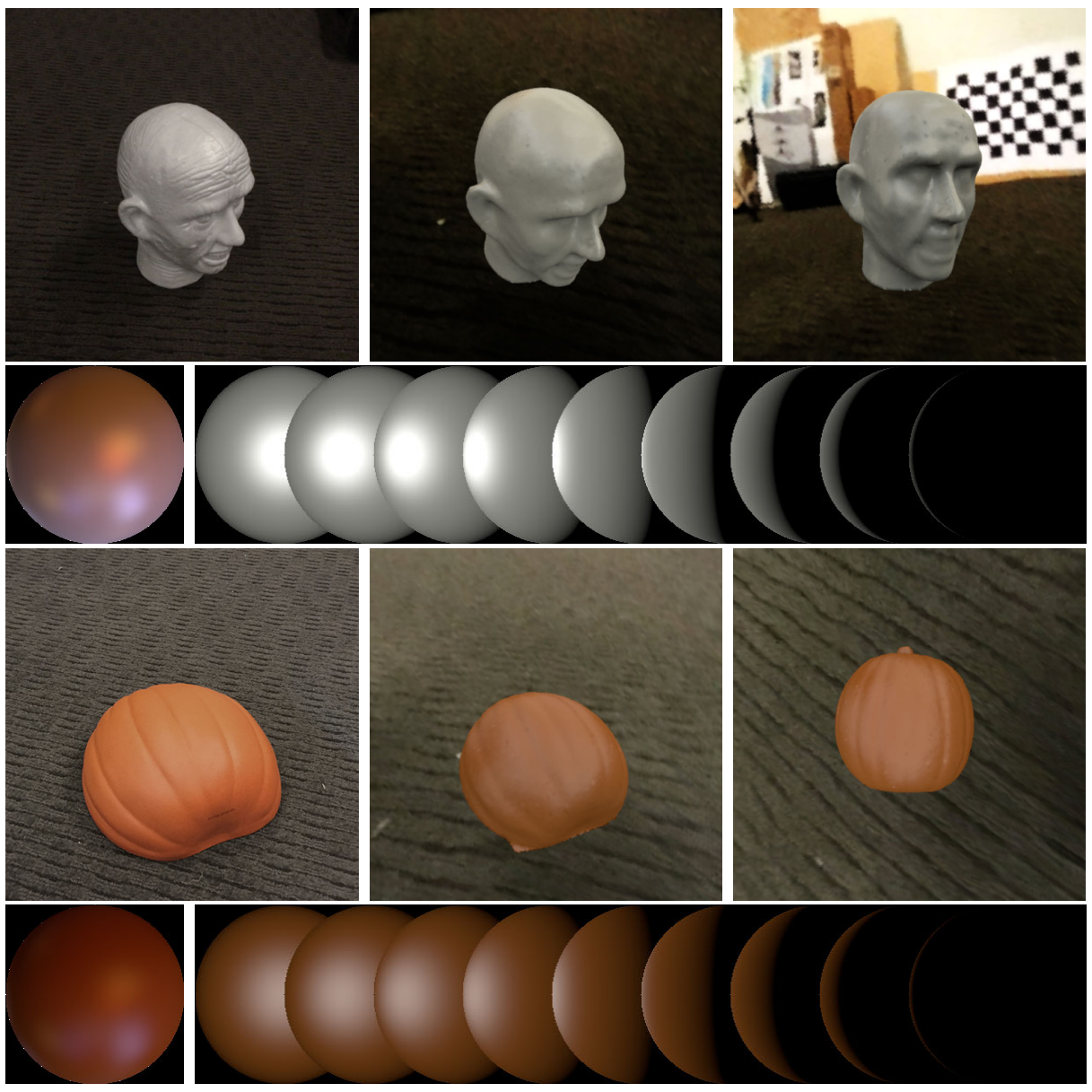

Head and Pumpkin models without a regularization (Eq. 4).

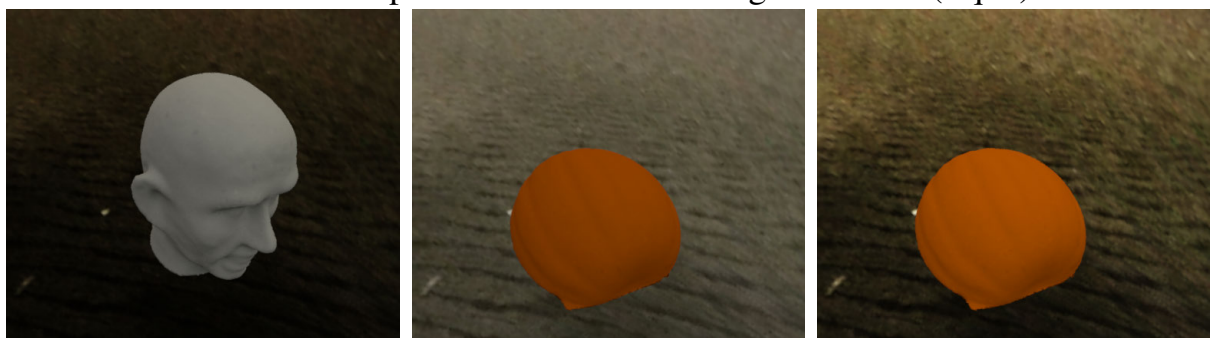

Figure 8. **Real data evaluation:** For each example, top-left is the input (real) scene, top-middle shows the rendered scene with estimated BRDF parameters, and top-right shows a different rendered view of the same scene. Bottom left is a rendered sphere with the estimated BRDF, and bottom-right is the rendered sphere with varying point light. Grouplet and HemiCNN with CubeRoot+$E_c$ were used for the first and second example, respectively.

## 4.3. Results on Real Data

Earlier in Sec. 1, and Sec. 3.2, we discussed about the potential issues of scale ambiguity that we might face in real-world data because of the fact that we aims to use commodity RGBD camera rather than HDR videos. As we expected, the regularization (Eq. 4) plays an important role for the correct results (Fig. 8). Notice that the result with the regularization captures more proper brightness as well as plausible gloss. The more views of the real examples are shown in the supplementary video.

## 5. Conclusions and Limitations

In this paper, we proposed a lightweight and practical approach for surface reflectance estimation directly from 8-bit RGB images in real-time, which can be plugged into 3D scanning-and-fusion systems with a commodity RGBD sensor for scene capture. Our approach is learning-based, with the inference time less than 90ms per material and model size less than 340K bytes. Compared to prior work, our method is a more feasible solution for real-time applications (VR/AR) on mobile devices. We proposed two novel network architectures, HemiCNN and Grouplet, to handle the unstructured measured data from input images. We also designed a novel loss function that is both perceptually-based and able to resolve the scale ambiguity and color-constancy ambiguity for reflectance estimation. In addition, we also provided the first large-scale synthetic data set (SynBRDF) as a benchmark dataset for training and evaluation for surface reflectance estimation in uncontrolled environments.

Our method has several limitations that we plan to address in future work. First, our method estimates homogeneous reflectance currently – while GroupLet and HemiCNN in theory can operate for each voxel separately and thus is able to estimate spatially-varying reflectance, in practice we found using more voxels as input often results in more robust estimation. One future direction is to jointly learn several basis reflectance functions and weight maps to estimate spatially-varying BRDF. Second, we use the isotropic Ward model for BRDF representation. In the future, we plan to investigate more general, data-driven models such as DSBRDF [25] and the related perceptually-based loss [34]. Finally, we are interested in using neural networks to jointly refine both 3D geometry and reflectance estimation, and using domain adaption techniques to further improve the performance on real data.

## Appendix 1

**Derivation of Eq.(4)** For viewing direction $\omega_o$, the observed scene radiance $L_o$ is given by

$$L_o = \int_{\omega_i} f(\omega_i, \omega_o; \Theta) \cdot L_i \cdot \max(\cos\theta_i, 0) d\omega_i, \quad (5)$$

where $L_i$ is the environmental illumination in the direction $\omega_i$. We simplified the above rendering equation so that all terms can be computed from the input fed into the networks. Suppose the environment illumination is uniform, i.e., $L_i = \bar{L}$, by integrating the reflected radiance from the entire hemisphere, the measured radiance is:

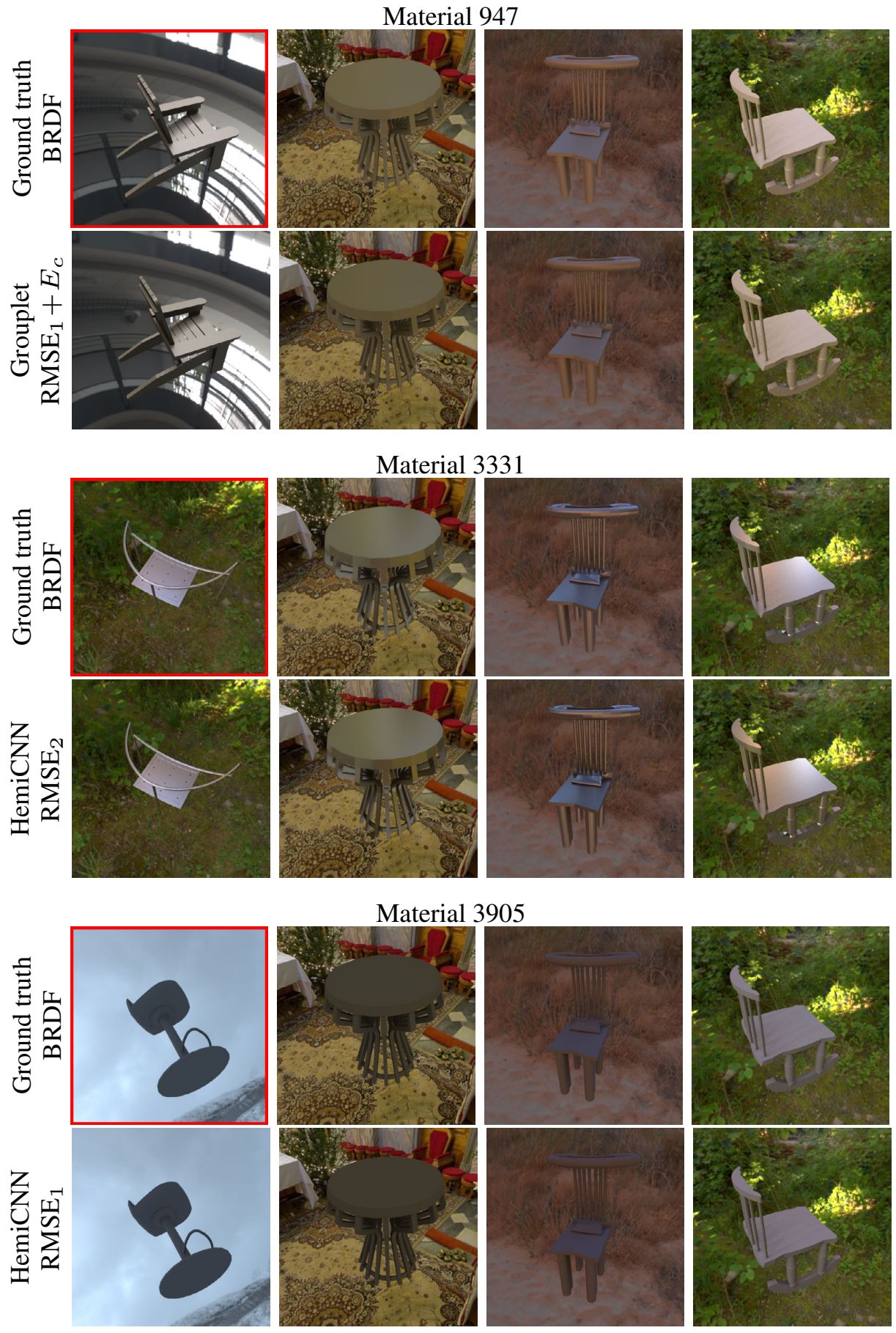

Figure 9. Qualitative results of three material examples. For each material, images in first row are rendered from ground truth BRDFs, and images in next the row show the rendered images from the estimated BRDF using one of the methods from the list in Table 2. The red-banded ground truth images are from the synthetic image sequences used for the inference (inputs). To demonstrate, even with the same BRDF, how the different objects and environmental lights could change the appearance of the scene, the three images from second to fourth columns are rendered from the same BRDF of each row. Notice how close the colors of surfaces and shadows in estimated scene in 947 and 3905 and glossiness on the chair and table in 3331, under different lighting to the ground truth. More examples from different set ups and losses from the list are in the supplementary material.

$$\bar{L_o} \approx (\rho_d + \rho_s)\bar{L}. \quad (6)$$

Both $\bar{L_o}$ and $\bar{L}$ can be approximated from input images, where the average intensity of the foreground object is close to $\bar{L_o}$, and the average intensity of the background is close to $\bar{L}$. Since the input images are 8-bit images in the sRGB color space rather than linear HDR images, we need to apply an additional gamma transformation between pixel intensities and scene radiance ($\gamma = 2.4$ for sRGB). Thus, we have $\bar{L_o} \approx \bar{F}^\gamma$ and $\bar{L} \approx \bar{B}^\gamma$, where $\bar{F}$ and $\bar{B}$ are the average image intensities for the foreground and background. Putting all together, we have

$$\bar{F}_i^\gamma \approx (\rho_d + \rho_s) \cdot \bar{B}_i^\gamma, \quad (7)$$

and thus we have the $E_c$ term in Eq.(4).

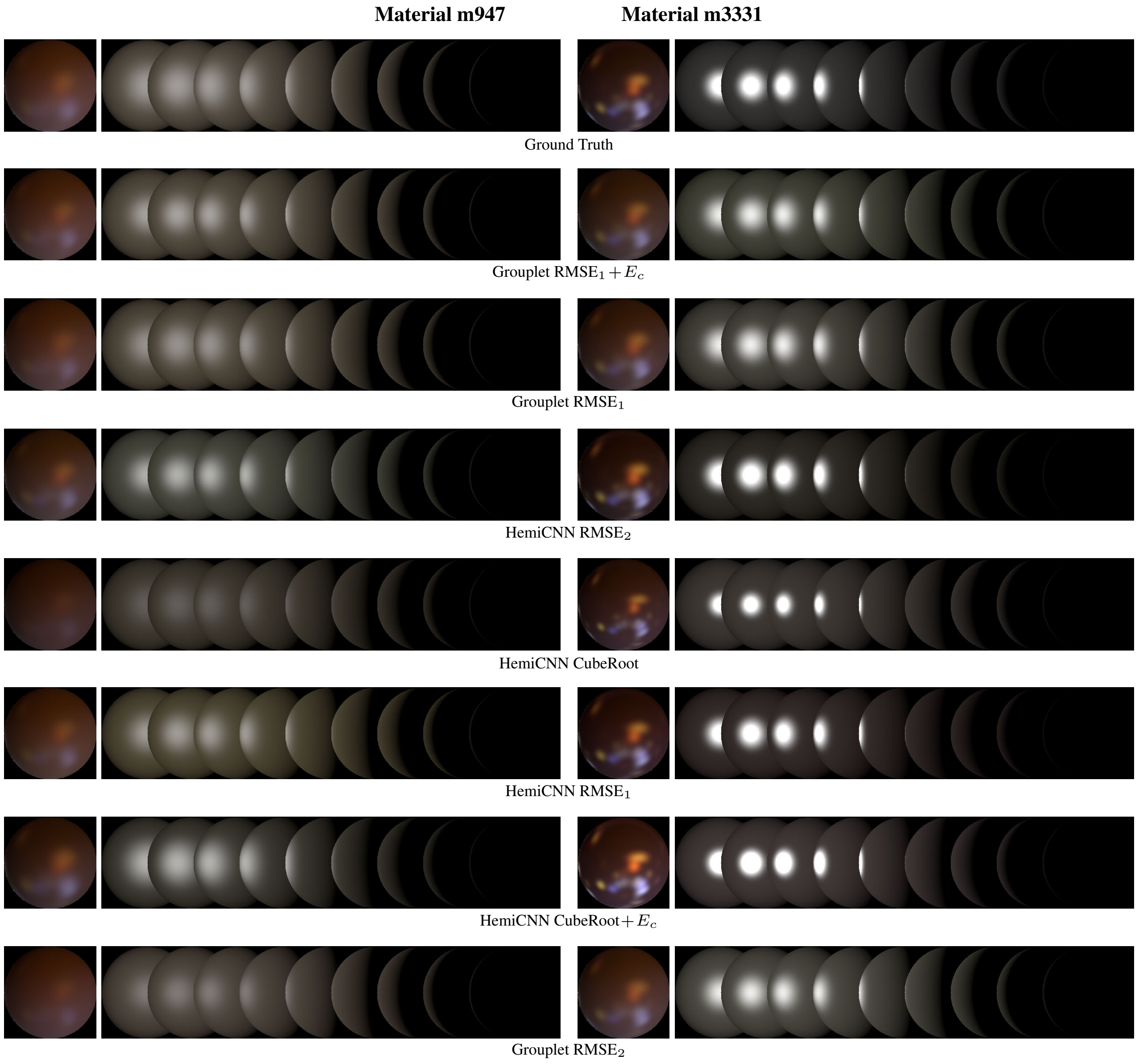

Figure 10. Qualitative comparison of proposed methods on *m947* (left) and *m3331* (right). For each material, we show a rendered image under natural illumination, as well as under varying light sources. The top row is the ground truth, while the remaining rows are the estimated results from variations of the proposed methods. Please refer Table 1 and the main paper for details.

| Network | Loss | RMSE | | | | | | | | | | |
|---|---|---|---|---|---|---|---|---|---|---|---|---|
| | | Avg | m3331 | m3905 | m1161 | m3522 | m3665 | m1891 | m586 | m2646 | m3601 | m947 |
| Grouplet | $\text{RMSE}_1 + E_c$ | 0.455 | 0.389 | 0.224 | 0.388 | 0.885 | 0.658 | 0.504 | 0.427 | 0.112 | 0.135 | 0.272 |
| Grouplet | $\text{RMSE}_1$ | 0.432 | 0.493 | 0.232 | 0.235 | 0.734 | 0.729 | 0.514 | 0.452 | 0.175 | 0.119 | 0.200 |
| HemiCNN | $\text{RMSE}_2$ | 0.564 | 0.153 | 0.130 | 1.341 | 1.012 | 0.588 | 0.480 | 0.406 | 0.026 | 0.184 | 0.370 |
| HemiCNN | CubeRoot | 0.439 | 0.356 | 0.095 | 1.282 | 0.841 | 0.680 | 0.197 | 0.557 | 0.233 | 0.304 | 0.536 |
| HemiCNN | $\text{RMSE}_1$ | 0.419 | 0.102 | 0.080 | 1.082 | 0.966 | 0.630 | 0.288 | 0.507 | 0.075 | 0.207 | 0.325 |
| HemiCNN | $\text{CubeRoot} + E_c$ | 0.583 | 0.481 | 0.161 | 1.274 | 0.878 | 0.742 | 0.269 | 0.590 | 0.283 | 0.330 | 0.454 |
| Grouplet | $\text{RMSE}_2$ | 0.457 | 0.493 | 0.232 | 0.440 | 0.707 | 0.665 | 0.514 | 0.378 | 0.058 | 0.155 | 0.386 |

Table 3. **Extension of Table 2** RMSE (w.r.t ground truth BRDF parameters normalized to zxero mean and unit variance) on the testing set of SynBRDF with the order of the user preference ranks from a perceptual study of ten randomly sampled materials. In addition to the average RMSE shown in the main paper, we add the RMSE for each of the ten materials.

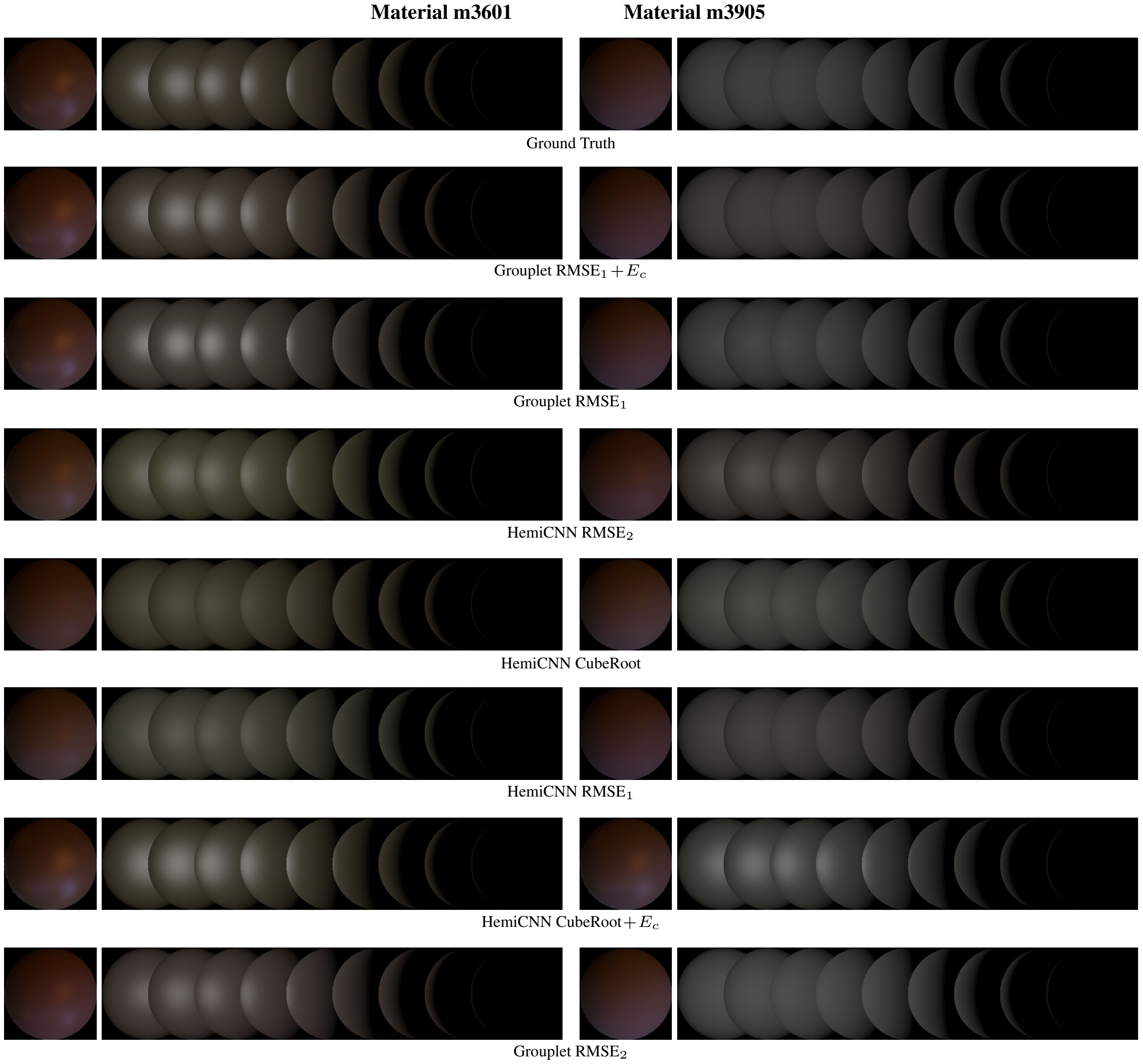

Figure 11. Qualitative comparison of proposed methods on *m3601* (left) and *m3905* (right): Please refer to Table 1.

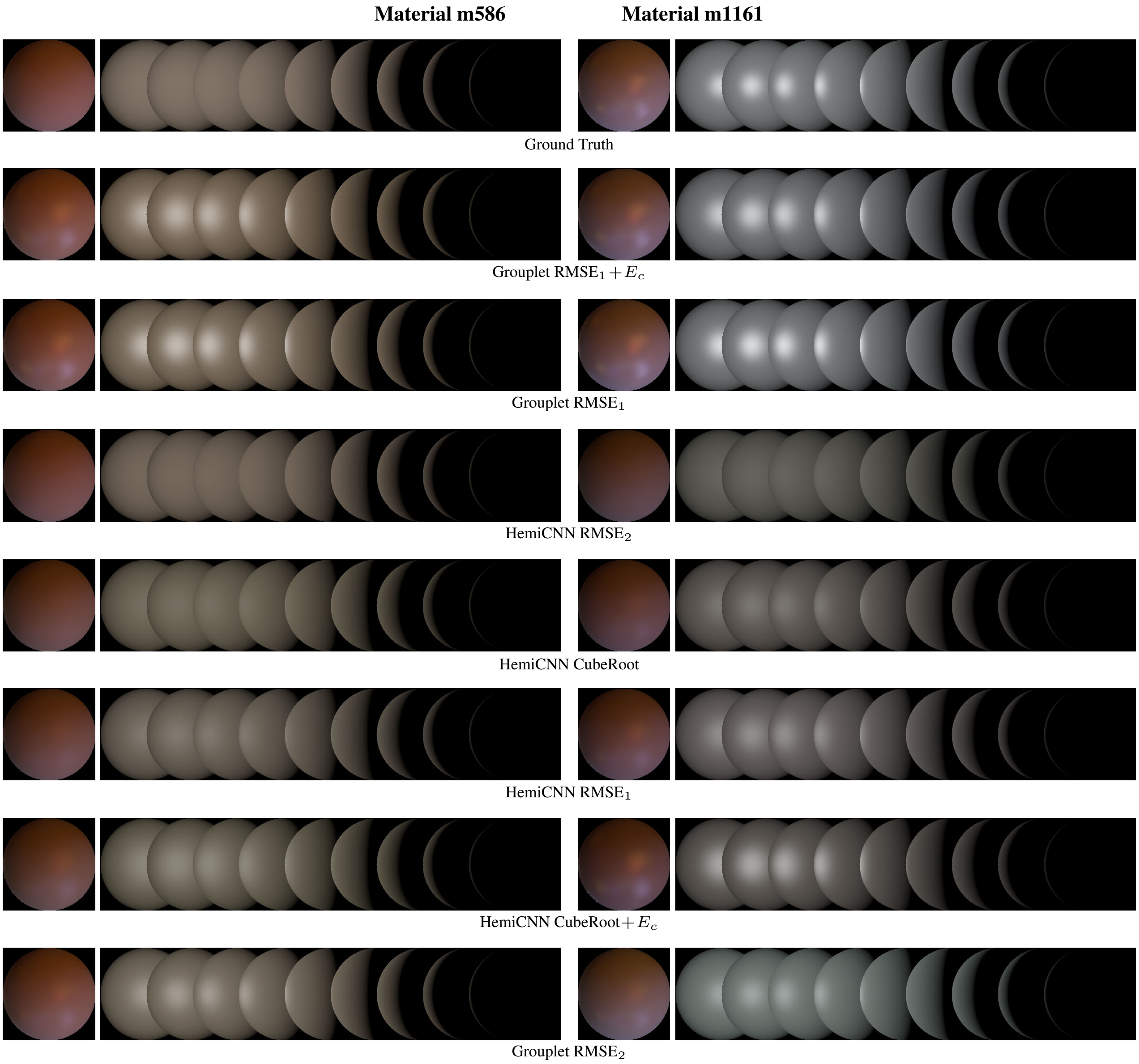

Figure 12. Qualitative comparison of methods for *m586* (left) and *m1161* (right): Please refer to Table 1.

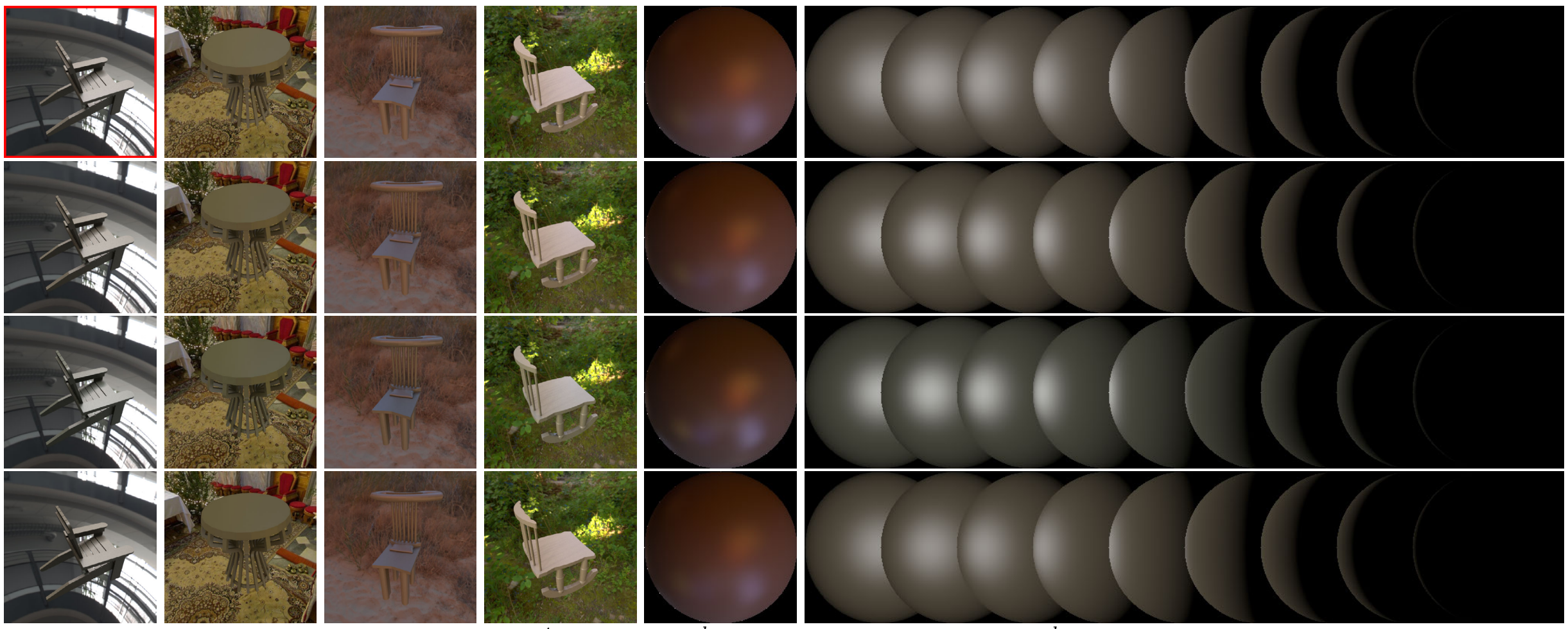

m947: $1^{st}$ row: GT, $2^{nd}$ row: Grouplet+RMSE$_1 + E_c$, $3^{rd}$ row: HemiCNN+RMSE$_2$, $4^{th}$ row: Grouplet+RMSE$_1$

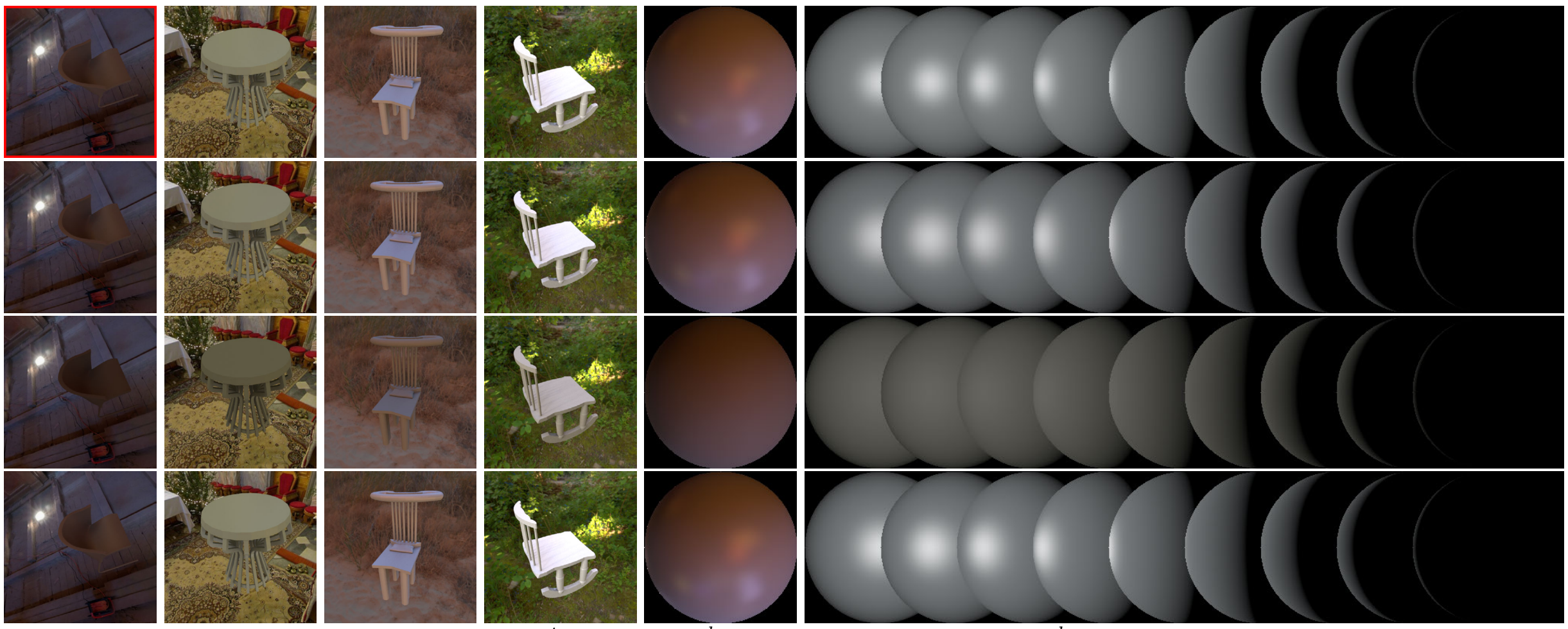

m1161: $1^{st}$ row: GT, $2^{nd}$ row: Grouplet+RMSE$_1 + E_c$, $3^{rd}$ row: HemiCNN+RMSE$_2$, $4^{th}$ row: Grouplet+RMSE$_1$

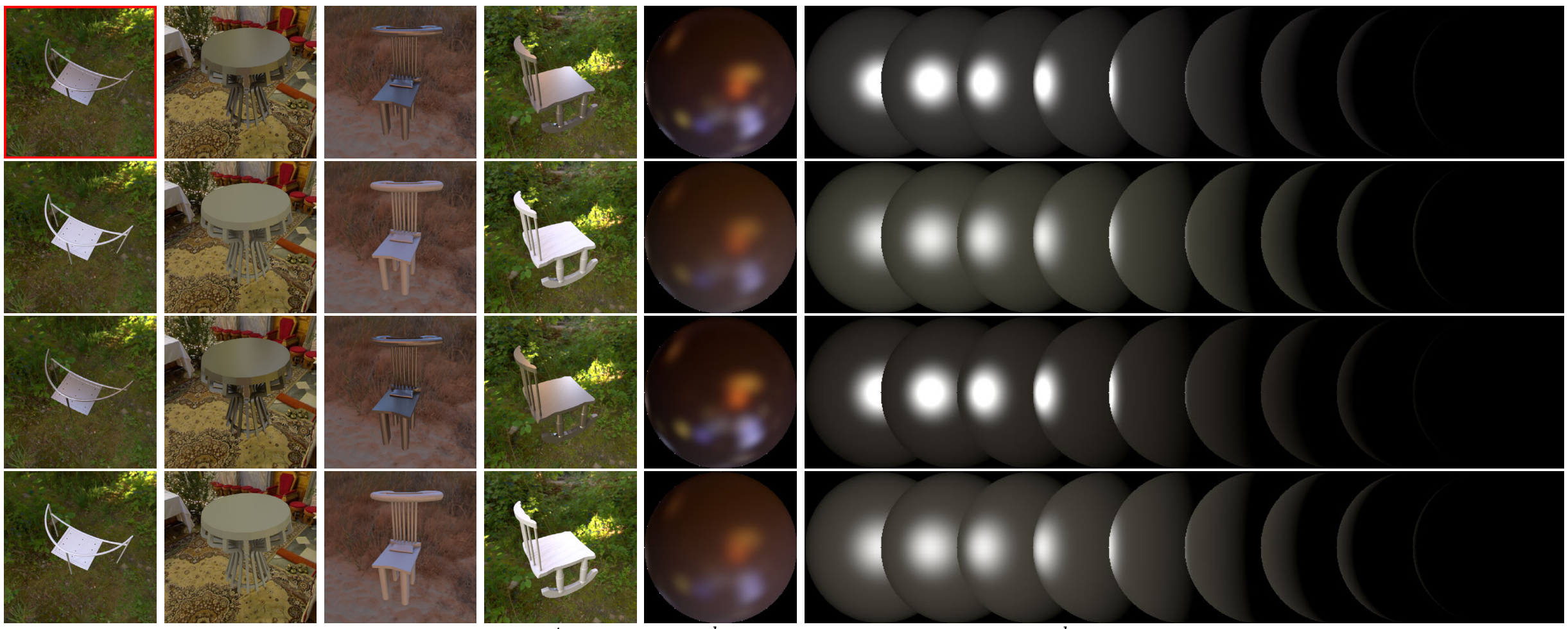

m3331: $1^{st}$ row: GT, $2^{nd}$ row: Grouplet+RMSE$_1 + E_c$, $3^{rd}$ row: HemiCNN+RMSE$_2$, $4^{th}$ row: Grouplet+RMSE$_1$

Figure 13. Additional qualitative evaluation. See Figure 9 in the main paper.